# PHONEME-LEVEL SPEECH AND NATURAL LANGUAGE INTEGRATION FOR AGGLUTINATIVE LANGUAGES *

Geunbae Lee
Jong-Hyeok Lee
Kyunghee Kim
Department of Computer Science & Engineering
and Postech Information Research Laboratory
Pohang University of Science & Technology
San 31, Hoja-Dong, Pohang, 790-784, Korea
gblee@vision.postech.ac.kr

ABSTRACT

A new tightly coupled speech and natural language integration model is presented for a TDNN-based large vocabulary continuous speech recognition system. Unlike the popular n-best techniques developed for integrating mainly HMM-based speech and natural language systems in word level, which is obviously inadequate for the morphologically complex agglutinative languages, our model constructs a spoken language system based on the phoneme-level integration. The TDNN-CYK spoken language architecture is designed and implemented using the TDNN-based diphone recognition module integrated with the table-driven phonological/morphological co-analysis. Our integration model provides a seamless integration of speech and natural language for connectionist speech recognition systems especially for morphologically complex languages such as Korean. Our experiment results show that the speaker-dependent continuous Eojeol (word) recognition can be integrated with the morphological analysis with over 80% morphological analysis success rate directly from the speech input for the middle-level vocabularies.

## 1 INTRODUCTION

A spoken natural language system requires many different levels of knowledge sources including acoustic-phonetic, phonological, morphological, syntactic, and semantic levels. The knowledge sources are grouped and processed in either speech processing models or statistical/symbolic natural language processing models. Since the speech and the natural language communities have conducted almost independent researches, these models were not completely integrated and often biased by neglecting either the acoustic-phonetic or the high-level linguistic information. The spoken language system requires seamless integration of speech signals into the high level language processing components. Recent advances in large vocabulary continuous speech recognition makes an integrated speech and natural language system possible and feasible. In a spoken language architecture, we must consider all the acoustic-

*This research was supported in part by a grant from KOSEF (Korean Science and Engineering Foundation). We also thank to WonIl Lee for coding the lexicon and the morphological parser and to professor Hong Jeong for his valuable suggestions for the earlier draft of this paper. An extended version of this paper was submitted to the journal of natural language engineering for a review.

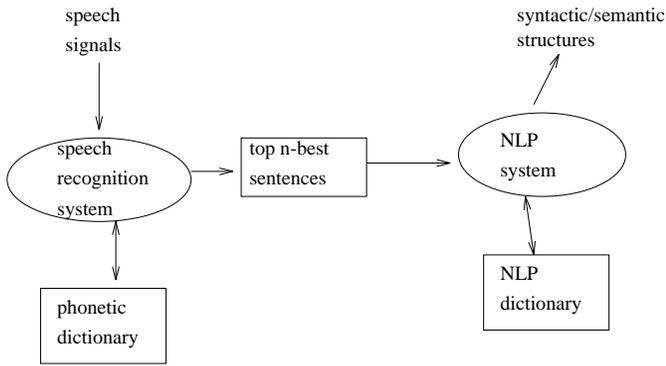

Figure 1: N-best search: current speech and natural language integration method

phonetic and linguistic information equally and choose the most feasible candidates at each acoustic and language processing step.

Current speech and natural language integration mainly relies on the word-level n-best search techniques [1] which are obviously inefficient for morphologically complex agglutinative languages such as Korean. Figure 1 shows the current n-best integration method.

For HMM-based speech recognition systems, the n-best search techniques [1, 2] have been successfully applied to the integration of speech recognition systems into the natural language systems. However, current implementations of the n-best techniques only support integration at the word level (using word sequences or lattice), and mainly used for the integration of existing speech and natural language systems [3, 4]. Also the n-best search is viable only for short sentences since the n grows exponentially with the sentence length (number of words in the sentence). Because the n-best search integrates at the word level, the natural language systems usually support word-level dictionary which seems to be a reasonable assumption in morphologically simple languages such as English. However, most natural language systems which deal with the morphologically complex languages currently use the morpheme-level dictionary for the linguistic generality. For these languages, the dictionary size for large vocabulary continuous spoken language system will grow very fast if we adhere to the full word-level phonetic dictionary because new words can be almost freely generated by concatenating the constituent morphemes (e.g. noun + postposition or verb + verb-endings in Korean). To incorporate the general morpheme-level dictionary into the spoken language system, we must develop a sub-word level integration technique between speech and natural language. The technique is more important in the languages which have very complex morphological structures caused by complex postpositions and verb-endings, such as Korean.

In this paper, we present a new integration architecture of speech and natural language based on the table-driven phonological/morphological co-analysis using the well-known dynamic programming technique [5] and the connectionist diphone spotting technique. Our model integrates a phonological/morphological parsing into a speech recognition, not at a word-level, but at a phoneme-level for a more tightly coupled integrated system. We present a new integration architecture, not for the popular HMM-based systems, but for recently developed connectionist speech recognition systems. Connectionist speech recognition [6] has several advantages compared with the classical symbolic and stochastic modeling. Especially, the time-delay neural network (TDNN) model [7] has been widely used to model the time shift invariance of speech signals. However, the integrated speech and natural language processing models using the TDNN have not been much researched before[1]. In this regard, we present a phoneme-level integration method for large vocabulary connectionist speech recognition model using the TDNN, especially for the morphologically complex agglutinative languages.

---

[1] One notable exception is the researches by Sawai [8, 9].

## 2 FEATURES OF SPOKEN KOREAN

Korean, which can be classified into a morphologically agglutinative and syntactically SOV languages, has several unique linguistic features. The followings are morphological and phonological features of spoken Korean for the understanding of our integration method. For the syntactic level features, [10] explains some Korean syntax modeling. In this paper, the Yale romanization is used for representing the Korean phonemes.

1) A Korean word, called Eojeol, consists of more than one morphemes with clear-cut boundaries in between. For example, an Eojeol *pha-il+tul+ul (files [obj])* consists of 3 morphemes:

pha-il (file) + tul (plural suffix) + ul (object case-marker)

2) Korean is a postpositional language with noun-endings, verb-endings, and pre-final verb-endings. These functional morphemes determine the noun's case roles, verb's tenses, modals, and modification relations between Eojeols. For example, in *swu-ceng-ha+yess+ten pha-il (the file that was edited)*, the verb *swu-ceng-ha (edit)* is of past tense and modifies *pha-il (file)* according to the given verb-endings:

swu-ceng-ha (edit) + yess (past tense prefinal verb-ending) + ten (adnominal verb-ending)

3) The unit of pause in a spoken Korean (called Eonjeol) may be different from that in a written Korean (called Eojeol). For example, in speaking *nay-ka e-cey swu-ceng-ha-yess-ten pha-il-tul-ul /tmp lo pok-sa-ha-ye-la* (spaces delimit Eojeols, meaning that "copy the files that I edited yesterday to /tmp"), a person may pause after *nay-ka* and after *e-cey swu-ceng-ha-yess-ten pha-il-tul-ul*, and after */tmp lo pok-sa-ha-ye-la*.

4) Phonological changes occur in a morpheme, between morphemes in an Eojeol, and between Eojeols in an Eonjeol. These changes include assimilation, dissimilation, contraction, and insertion. For example, a morpheme *pok-sa* is pronounced as *pok-ssa* (dissimilation, meaning "copy"), and *kwuk-min* is pronounced as *kwung-min* (assimilation, meaning "nationality"). An Eojeol *su-ceng-ha-yess-ten* is pronounced as *su-ceng-ha-yet-tten*.

## 3 SYSTEM ARCHITECTURE FOR SPEECH AND NATURAL LANGUAGE INTEGRATION

Our integration technique employs a phoneme lattice and a morpheme-level phonetic dictionary. This can be more microscopic integration compared with the classical approaches of using the word lattice and the word-level dictionary, such as the n-best integration technique which is mainly used for English. The phoneme lattice makes the phonological rule modeling possible in an early stage of spoken language processing. The phonological/morphological analysis can be performed together using the morpheme-level phonetic dictionary, and the dictionary size becomes stable regardless of the vocabulary size because new vocabularies can be generated by combining existing morphemes in the dictionary. Unlike the conventional integration method which uses the separate dictionaries for the speech recognition and the natural language processing, our integration model uses a unified morpheme-level phonetic dictionary together with the declarative morphotactic and phonotactic information. In our spoken language architecture, we employ a hierarchy of diphone spotting TDNNs for the acoustic-level processing, and develop a phonological/morphological co-analysis technique for the seamless integration. The output of the integrated architecture can be directly fed to the conventional natural language syntax/semantics analysis systems. Figure 2 shows the integrated spoken language pro-cessing architecture, a tightly coupled integration model of speech and natural language. The speech signal is analyzed using the TDNN diphone recognizer. The diphone recognizer also rearranges the diphone strings to produce the phoneme lattice. From the phoneme lattice, the morphological analyzer produces the morphologically analyzed Eojeols by handling the morphological segmentation, morphotactics verification, and the irregular conjugation. The phonological processing is integrated into the morphological parsing through the declarative phonological rule modeling. In the next section, we will explain the speech recognition and the morphological/phonological processing in detail.

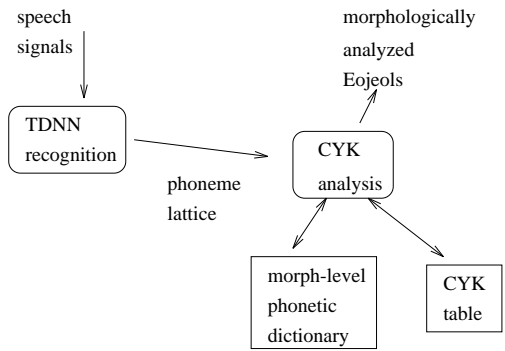

Figure 2: TDNN-CYK integration architecture

| diphone groups | diphone numbers |
|---|---|
| V | 21 |
| C1V | 378 |
| VC2 | 147 |
| C2C1 | 126 |

Figure 3: Korean diphone groups (V: vowel, C1: syllable-first consonant, C2: syllable-final consonant)

## 4 Diphone-based speech recognition

For large-vocabulary continuos speech recognition, the sub-word level recognition must be supported. We selected a group of diphones for the sub-word unit because direct phoneme recognition in Korean is very difficult. The 46 Korean phonemes are very similar each other especially in the following cases: 1) the Korean diphthongs are hard to distinguish from the mono-vowels, and 2) the syllable-final consonants are hard to differentiate from the syllable-first consonants. The selected diphone groups (figure 3) have more information than the phonemes and are much fewer in numbers than the popular triphones [11].

Figure 4 shows the diphone-based TDNN speech recognition system. The system consists of total 19 different TDNN networks for recognition of the Korean diphone groups.

The speech recognition is performed through the following steps (for more details, see [12]):

1) Pre-processing: The digitized speech signal is segmented into 200 msec size, 512 order FFTed and 16 step mel-scaled to obtain the filter-bank coefficients. For the endpoint detection, the short-time energy and the zero-crossing rate are used. Each frame size is 10 msec and the 20 frames of 16 value normalized filter-bank coefficients are fed to the vowel group recognition TDNN.

2) Vowel group recognition: The input is 20 frame vectors (20*16 = 320 units) and the output is 18 units for the 18 vowel groups (the 17 groups according to the contained vowels and one CC group with no vowel). For each vowel group, separate diphone recognition TDNN is invoked, and the system has a hierarchical TDNN architecture. Each TDNN has the standard architecture which is well described in [7].

3) Diphone recognition: According to the

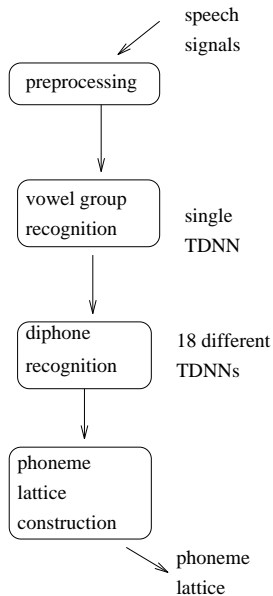

Figure 4: Diphone-based TDNN speech recognition system

recognized vowel group, each pertinent diphone recognition TDNN is activated. For each TDNN, the input is the same 20 frame vectors, and the output is the classified diphones for each vowel group. For example, for the /ya/ vowel group, there are total 15 output units: 9 for C1V type diphones (/kya/, /nya/, /tya/, /lya/, /mya/, /pya/, /sya/, /kkya/, /hya/), 5 for VC2 type diphones (/yak/, /yal/, /yan/, /yam/, /yang/) and one for V type diphone (/ya/). Each of the 18 TDNNs has the different number of output units according to the number of diphones in each vowel group.

4) Diphone2phoneme decoding: From the resulting diphone sequences, this step obtains the phoneme lattice which contains the candidate phoneme sequences. We use a simple deterministic decoding heuristics without any probabilistic calculations, and try to maintain all the possible diphone spotting results in the phoneme lattice since the later phonological/morphological processing can safely prune the incorrect recognitions. The decoding begins by grouping the diphones into the same types (C1V, V, VC2, C2C1 types). The frequency count for each diphone, that is, the number of specific diphones per 10 msec frame shift, is utilized to fix the insertion errors by deleting the lower frequency count diphones, and finally the diphones are split into the constituent phonemes by merging the same phonemes in the neighboring diphones. This simple non-probablistic decoding scheme surprisingly works well for our domain, and the resulting phoneme lattice reliably provides all the possible output phonemes in the speech recognition.

## 5 Morphological analysis from the phoneme lattice

The morphological analysis transforms the phoneme lattice into the sequences of morphologically analyzed Eojeols (which is a unit of spacing in Korean orthography and usually consists of single noun or verb-stem plus several functional morphemes). Our morphological analysis takes a phoneme lattice rather than a phoneme string as an input since we want to have a chance to exploit all the speech recognition results during the morphological analysis. The phoneme lattice provides alternative phonetic transcriptions of speech sounds which must be transformed to produce the orthographic morpheme strings. Unlike the conventional morphological analysis from the written text input, the morphological analysis of the *phoneme lattice* must solve the following subproblems: 1) The phonetic transcriptions must be segmented and mapped into the orthographic morphemes which are basic units of written language processing. 2) The phonological changes that can be captured by the Korean phonological rules must be modeled and processed during the morphological analysis. 3) An efficient dictionary search is required because the phoneme lattice results in exponential number of phoneme chains.

The Korean morphological analyzer [13] was implemented based on the well-known CYK parsing technique [5] and augmented in order to handle the Korean phonological changes and phoneme lattice input. Figure 5

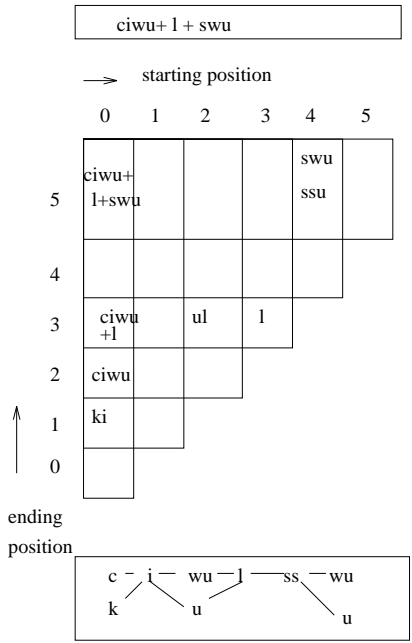

Figure 5: Morphological parsing of the phoneme lattice (from top: morphologically analyzed output Eojeol, CYK triangular table, input phoneme lattice). The example phoneme lattice was obtained from the input speech *ci-wul-ssu* (deletable) using the diphone-based TDNN speech recognition system, and the morphological analysis produces *ci-wu+l+su*, where "+" is the morpheme boundary, and "-" is the syllable boundary. The CYK triangular table was filled in with all the possible morphemes which are obtainable from the dictionary look-up, and also with all the possible morpheme combinations.

shows our morphological analysis scheme for the phoneme lattice.

The basic process of the Korean morphological analysis consists of the morpheme segmentation, checking the possible morpheme connectivity (handling of the morphotactics), and the reconstruction of the original morphemes from the irregular conjugations (handling of the orthographic rules).

The morpheme segmentation is performed using the morpheme entry in the dictionary. During the left to right scanning of the input text, when the morpheme is found in the dictionary, it is enrolled in the CYK table in the proper position. For example, in figure 5, the 3 different morphemes, that is, verb *ci-wu*, adnominalizing verb-ending *l*, and the bound noun *swu* are enrolled in the position (0,2), (3,3), and (4,5) respectively. The position (i,j) designates the start and end position of the input characters, and the verb *ciwu* starts in the position 0 (first position) and ends in the position 2, hence consists of 3 characters. We enroll all the matched morphemes on the input string in the CYK table (see figure 5 for other possible morphemes). During the segmentation, the possible morpheme connectivity must be checked for the selection of the correct morpheme boundaries for the input string. The morpheme connectivity can be verified from the Korean morphotactic information. The morphotactic information is included in the dictionary using the specialized Korean part-of-speech symbols (called connectivity information) [13]. We divided the major 13 Korean part-of-speech symbols into about 200 different refined symbols (tags) for the efficient verification of the connectivity of each morpheme, and constructed the morpheme connectivity matrix which designates the possible relative placement of 200 refined part-of-speech tags in the string. For example, in figure 5, the morpheme *ci-wu* (verb stem, meaning "delete") can be in the left side of the morpheme *l* (adnominalizing verb-ending) because the morpheme connectivity matrix verifies that the connection of *verb stem* to the *adnominalizing verb-ending* is legal. The CYK table provides the possible positions of the connectivity checking. For example, in figure 5, the connectivity information of *ci-wu* and *l* is worth checking because the position (0,2) and (3,3) can be concatenated to produce the position (0,3) so the result *ci-wu+l* is put in the position (0,3). The irregular conjugations are handled in declarative way by putting the inflected forms as well as the original forms of the morphemes in the dictionary.

The above-mentioned basic morphological analysis scheme was augmented to solve the three sub-problems in handling the phoneme lattice input and the phonological changes during the morphological analysis.

| phonetic transcription header | original morpheme | left morphological connectivity | right morphological connectivity | left phonemic connectivity | right phonemic connectivity |
|---|---|---|---|---|---|
| ci-wu | ci-wu | regular verb | regular verb | 'c' sound no-change | 'wu' sound no-change |
| l | l | adnominalizing verb-ending | adnominalizing verb-ending | 'l' sound no-change | 'l' sound no-change |
| sswu | swu | bound-noun | bound-noun | 's' sound change to 'ss' | 'wu' sound no-change |

Figure 6: The morpheme-level phonetic dictionary. The figure shows three different morpheme entries *ci-wu, l, swu* with their phonetic transcription headers, orginial morphemes, left and right morphological connectivity, and left and right phonemic connectivity information. In the actual dictionary implementation, the morphological and phonemic connectivity information is encoded using the specialized symbols. The left and right distinction is for the morphemes that have the different connectivity information to the left concatenated and right concatenated morphemes.

1) For phonetic transcription into the orthographic morpheme mapping, we indexed each morpheme in the dictionary by the corresponding phonetic transcription header, and constructed so called morpheme-level phonetic dictionary. The single phonetic transcription can be associated with many different morpheme entries for the homophone style morphemes. In this way, the accessing of phonetic headers can lead to all the corresponding morphemes in the orthographic forms. Figure 6 shows the morpheme-level phonetic dictionary.

2) In Korean, the phonological changes can occur within the morpheme or across the morpheme boundary. For the former case, the phonetic transcription headers in the dictionary already reflect the phonological changes since the dictionary entry is the whole morpheme. However, for the latter case, we have to model the Korean phonological rules to handle the between-morpheme-phonological changes. We declaratively modeled the major Korean phonological rules including the 2nd consonant standardization, consonant assimilation, palatalization, glottalization (consonant dissimilation), and insertion according to the Korean Ministry of Education Standard, and processed the Korean phonology during the morphotactic verification. The declarative phonological rule is encoded in the left and right phonemic connectivity information in the dictionary. For example, the bound noun *swu* has phonetic realization *sswu* after the *l* sound. Figure 6 designates the phenomenon in the left phonemic connectivity information in the *swu* entry. The separate phoneme connectivity matrix records all the possible relative phoneme placement much like the morpheme connectivity matrix. When the morphotactics is checked, the phoneme connectivity matrix is also checked to verify the possible phonological changes between the morphemes.

3) To handle the phoneme lattice search, we use the TRIE indexing for the fast dictionary access [14]. The breadth-first search on the TRIE structures for the phonetic transcription header can prune the unnecessary paths efficiently, and hence deal with the complexity of the phoneme lattice search.

## 6 Implementation and the experiment results

The TDNN-CYK spoken language system was implemented using C and standard X-window user interface under the UNIX/Sun Sparc platforms. The system's inputs are carefully articulated Korean speech in the normal laboratory environment, and the outputs are morphologically analyzed Eojeol sequences that can be directly fed to the conventional natural language syntax analysis system [10]. We constructed a 1000 entry morpheme-level phonetic dictionary in the UNIX operating system domain, and about over 100 entries of morpheme connectivity and phoneme connectivity matrix for the phonological/morphological analysis.

The dictionary is indexed using the phoneme-based TRIE to handle the phoneme lattice search. Since we don't have any standard segmented Korean speech database yet, we constructed our own by recording and manually segmenting 73 most frequent Korean diphones. The 73 diphones are acquired from the 300 Korean Eojeols (each Eojeol is pronounced 15 times by a female speaker) in the 100 Korean sentences, which can appear in the natural language commanding to the UNIX operating system[15].

Several experiments were performed to verify the system's performance of time-shift invariance, diphone recognition, and final Eojeol recognition including the morphological analysis. Belows are the brief results of each performance test. In each experiment, the input speech patterns are prepared as follows: Eojeols were recorded in a normal laboratory environment with an average S/N ratio of 12 dB. Speech data were sampled at 16kHz, and hamming windowed. From this windowed data, 512-point DTFTs were computed at 5 msec intervals. The DTFTs were used to generate 16 Mel-scale filter-bank coefficients at 10 msec intervals [7]. These spectra were normalized to produce suitable input levels for the four-layer structured TDNNs. We used hyperbolic arc tangent error function in the weight updating [16] in the back propagation training. We updated the weights after a small number of iterations [17].

### 6.1 Time-shift invariance of Korean diphones

We generated 2400 diphone samples for the typical 12 Korean diphones. The input patterns for the two tests are set the same in order to compare the *no shift* and *shift* cases. Figure 7 shows that the Korean diphone recognition has the time-invariance property of TDNN and suggests the optimal time interval near 200 - 250 msec for the diphones. These results imply that the context-independent diphone-based TDNN recognition is possible.

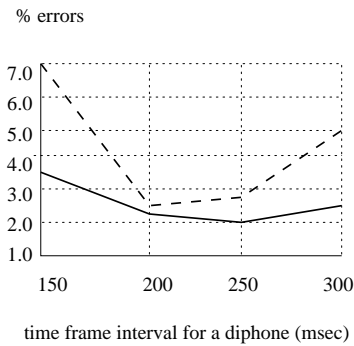

Figure 7: Average error rate of the segmented time frame (solid lines) versus the same time frame with maximum 40 msec left or right temporal shift (grey lines)

| unit of recognition | number of targets | number of samples | recognition rate |
|---|---|---|---|
| phoneme | 9 | 1080 | 94.06% |
|  | 17 | 2040 | 89.80% |
| diphone | 9 | 1080 | 95.42% |
|  | 17 | 2040 | 95.27% |

Figure 8: Diphone recognition versus phoneme recognition test

### 6.2 Comparison of diphone recognition vs. phoneme recognition

This experiment is to show that the diphone can improve the recognition rate of the Korean vowels regardless of many rising diphthongs, compared with the phoneme recognition. In the test, we set 150 msec time range for the phoneme and 200 msec for the diphone segmentation. Compared with the phoneme recognition case, figure 8 shows that the recognition rate of diphones doesn't decrease much when the number of targets with similar features doubly increases. Moreover, the unit with more than one feature can be efficiently recognized at the high rate in the diphone recognition.

a. continous diphones

| | total | correct | delete | insert |
|---|---|---|---|---|
| pattern size (rec. rate) | 7772 | 7259 (93.4%) | 513 (6.6%) | 3000 (38.6%) |

b. segmented diphones

| | vowel group | sub-TDNNs average | total average |
|---|---|---|---|
| rec. rate | 94.8% | 98.2% | 93.8% |

Figure 9: Continuous diphone spotting versus segmented diphone spotting

## 6.3 Performance of continuous diphone recognition

In this experiment, we extracted most typical 72 diphones in Korean from 66 Eojeols, each of which is pronounced 15 times to generate about 5500 diphone patterns. The 5500 training samples are used to train the vowel group TDNN and 10 different sub-TDNNs for each diphone group. During the recognition, the new 262 Eojeols are selected to generate the test patterns of 2432 Eojeols, and shifted 30 msec during the application to obtain the TDNNs diphone spotting performances in a continuous speech. Figure 9-a shows the continuous diphone spotting performance. We have total 7772 target diphones from the 2432 test Eojeol patterns. The *correct* designates that the correct target diphones were spotted in the testing position, and the *delete* designates the other case. The *insert* designates that the non-target diphones were spotted in the testing position. To compare the ability of handling the continuous speech, we also tested the diphone spotting using the hand segmented test patterns with the same 7772 target diphones. Figure 9-b shows the segmented diphone spotting performance. Since the test data are already hand-segmented before input, there are no insertion and deletion errors in this case. The fact that the segmented speech performance is not much better than the continuous one (93.8% vs. 93.4%) demonstrates the diphone's suitability to handling the continuous speech.

## 6.4 Performance of continuous Eojeol recognition

In order to test the ability of the full Eojeol recognition including the phoneme decoding and morphological analysis performance, a middle-vocabulary experiment was carried out. The task is a speaker-dependent and continuous Eojeol recognition which produces the morphologically analyzed Eojeol sequences. In the process, the speech recognizer produces the phoneme lattice that includes the correct phoneme sequence in the input Eojeol, and then the morphological analyzer produces the analyzed Eojeol from the phoneme lattice. So, in this task, all the intermediate steps, that is, diphone spotting, phoneme lattice decoding and morphological analysis from the phoneme lattice, are combined to produce the final recognition performance. The same 262 Eojeols in section 6.3 are fed to the total integrated system that has the pre-trained TDNN networks. Figure 10 shows the final performance of the continuous Eojeols. We have total 9605 target morphemes from the same 2432 test Eojeol patterns used in section 6.3. In the figure, the *correct* designates that the correct morpheme sequences can be analyzed from the speech input, and the *delete* means that the correct morpheme sequences cannot be generated. The *insert* designates the percentage of the spurious morphemes that are generated from the insertion errors. The performance is above 80% in the final morphological analysis success rate, which is promising but still relatively low compared with the continuous diphone recognition. The relatively low performance is due to the large insertion errors during the long range of continuous speech which cannot be handled properly in the phoneme lattice decoding. However, the morphological analyzer performed perfectly when the phoneme lattice contains the correct phoneme sequences.

|  | total | correct | delete | insert |
|---|---|---|---|---|
| pattern size (rec. rate) | 9605 | 7696 (80.1%) | 1909 (19.8%) | 7182 (74.76%) |

Figure 10: Continuous Eojeol recognition including morphological analysis

## 7 Comparison with the related researches

Recently, the idea of sending only the best n speech recognition results to the natural language system has been implemented using the time-synchronous Viterbi-style beam search algorithm [1]. The algorithm was also improved by the word-dependent search [2] and by adding the A* backward tree search [18]. The n-best integration is mainly utilized for the HMM-based continuous speech recognition systems, and many existing speech systems and natural language systems were successfully integrated using the n-best word search techniques [3, 4]. However, until now, the n-best search techniques are only implemented to directly produce the n-best sentences using the word sequences or word lattice, and this word-level integration was successful for the morphologically simple languages such as English. On the contrary, our integration is at the phoneme-level using the phoneme lattice because we need more sophisticated phonological/morphological handling in the integration process. The word-level n-best integration also assumes the word-level dictionary which is an unreasonable assumption for the morphologically complex languages.

The HMM-LR integration [19, 20] was implemented using the HMM's phoneme spotting ability integrated with the generalized LR parsing techniques [21]. Unlike the n-best integration, the HMM-LR integration was more tight and implemented at the phoneme-level by extending the LR parser's terminal symbols to cover the phonetic transcriptions. In this scheme, the LR parsing selects the most probable parsing results by obtaining the probability of the end-point candidate phonemes from the HMM's forward probability calculation. So the total integrated system is working by the LR parser's prediction of the next phoneme candidates which are then verified by the HMM's phoneme spotting abilities. The idea of extending the LR grammar to the phonetic transcriptions seems to be working for the phoneme-level integration. However, the scheme doesn't have any separate language-level dictionary, which results in the degenerated phonological/morphological processing, and also has the difficulty in the necessary scale-ups. On the contrary, our TDNN-CYK integration focuses on the general phonological/morphological handling which is essential for the agglutinative languages.

The idea of extending LR grammar to the phonetic transcriptions was also applied to the TDNN-LR integration method [9, 8] which was similarly implemented by replacing HMM's phoneme spotting by the TDNN's phoneme spotting. The integration was implemented by dynamic time warping (DTW) level-building search [22] between TDNN's phoneme sequences and LR grammar's phoneme sequences. However, the performance was relatively poor compared with the HMM-LR integration method [9]. There are basically two reasons for the poor TDNN-LR performances compared with the HMM-LR integration: 1) the TDNN model has rarely been applied to the practical large vocabulary systems yet, therefore it lacks in the fine tuning compared with the popular HMM models, and 2) the TDNN model has yet to find a right way to be effectively integrated into the natural language processing model. The HMM model supports a natural integration into the general chart-based parsing models such as generalized LR parsing because there are well-defined probablistic search techniques to be integrated. However, output activations of the multiple TDNNs are difficult to normalize and therefore difficult to naturally integrate into the popular probabilistic search schemes such as Viterbi search. Our TDNN-CYK method doesn't employ any probabilistic search in its integration, but send the entire phoneme lattice to the morphological analyzer. In this way, we can exploit all the TDNN's outputs in the language processing level which is somewhat inefficient but safe for the current scheme.

## 8 CONCLUSION

This paper presents a phoneme level integration of speech and natural language in a connectionist speech recognition model for agglutinative languages such as Korean. Our model's main contribution is to define the phoneme level integration that can support sophisticated phonological/morphological processing in the integration of speech and language, which is essential for the morphologically complex agglutinative languages. Also, the TDNN-CYK integration is a first attempt to develop a morphologically general integration model using the connectionist speech recognition paradigm.

Our TDNN-CYK spoken language architecture has many novel features for speech and natural language processing. First, the diphone-based TDNN proposes a nice subword unit of recognition, well reflecting the Korean phonetic characteristics. Secondly, the morphological analysis combined with the declarative phonological rule modeling is well suited to the phonetic transcription into the orthographic morpheme mapping, which is an essential task for every spoken language processing model. Finally, the TRIE structured phonetic transcription indexing can serve to reduce the phoneme access complexity in the direct morphological analysis from the phoneme lattice.

The experiments show that the final Eojeol recognition is over 80% in the middle-vocabulary speaker-dependent continuous Eojeol recognition, which is very promising in considering the continuous speech and the combination of several steps of performances such as diphone spotting, phoneme lattice decoding and morphological analysis. However the performance is relatively low compared with the continuous diphone recognition (which is over 93% in the same condition) because of the enormous insertion errors for long duration speech (Eonjeol or phrase). To recover from the insertion errors, we plan to incorporate an error correcting scheme into our phoneme decoding process that will result in the error-free phoneme lattice from which the morphological analyzer can produce the perfect analysis results.

## REFERENCES

[1] Y. L. Chow and R. Schwartz, "The n-best algorithm: An efficient procedures for finding top N sentence hypothesis," in *Proceedings of the second DARPA workshop on speech and natural language*, Los Altos, CA, 1989, Morgan Kaufmann Publishers, Inc.

[2] R. Schwartz and S. Austin, "Efficient, high-performance algorithms for n-best search," in *Proceedings of the third DARPA workshop on speech and natural language*, Los Altos, CA, 1990, Morgan Kaufmann Publishers, Inc.

[3] M. Agnas, H. Alshawi, I. Bretan, D. Carter, K. Ceder, M. Collins, R. Crouch, V. Digalakis, B. Ekholm, B. Bamback, J. Kaja, J. Karlgren, B. Lyberg, P. Price, S. Pulman, M. Rayner, C. Samuelsson, and T. Svensson, "Spoken language trnaslator: first year report," Technical Report ISRN SICS-R-94/03-SE, Swedish Institute of Computer Science and SRI International, 1994.

[4] M. Bates, R. Bobrow, P. Fung, R. Ingria, F. Kubala, J. Makhoul, L. Nguyen, R. Schwartz, and D. Stallard, "The BBN/HARC spoken language understanding system," in *Proceedings of the ICASSP-93*, 1993.

[5] A. Aho and J. D. Ullman, *The theory of parsing, translation, and compiling, Vol 1: parsing*, Prentice-Hall, Englewood Cliffs, NJ, 1972.

[6] D. Morgan and C. L. Scofield, *Neural networks and speech processing*, Kluwer Academic Publishers, Inc., 1991.

[7] A. Waibel, T. Hanaazawa, G. Hinton, K. Shikano, and K. Lang, "Phoneme recognition using time-delay neural networks," *IEEE Transactions on Acoustics, Speech and Signal Processing*, vol. 37, no. 3, pp. 328 – 339, 1989.

[8] H. Sawai, "The TDNN-LR large-vocabulary and continuous speech recognition system," in *Proceedings of the international conference on spoken language processing (ICSLP)*, 1990.

[9] H. Sawai, "TDNN-LR continuous speech recognition system using adaptive incremental TDNN training," in *Proceedings of the ICASSP-91*, 1991.

[10] W. Lee, G. Lee, and J. Lee, "Table-driven neural syntactic analysis of spoken Korean," in *Proceedings of COLING-94*, 1994.

[11] K. F. Lee, *Automatic speech recognition*, Kluwer Academic Publishers, Inc., 1989.

[12] K. Kim, G. Lee, and J. Lee, "Integrating TDNN-based diphone recognition with table-driven morphology parsing for understanding of spoken Korean," in *Proceedings of the international conference on spoken language processing (ICSLP)*, 1994.

[13] E. C. Lee and J. H. Lee, "The implementation of Korean morphological analyzer using hierarchical symbolic connectivity information," in *Proceedigns of the 4th conference on Korean and Korean information processing*, 1992 (in Korean).

[14] A. V. Aho, J. E. Hopcroft, and J. D. Ullman, *Data structures and algorithms*, Addison-Wesley Publishing Company, 1983.

[15] W. Lee and G. Lee, "From natural language to shell-script: A case-based reasoning system for automatic unix programming," *Expert systems with applications: An international journal*, vol. 9, no. 2, , 1995 (in press).

[16] S. E. Fahlman, "Faster-learning variations on back-propagation: An empirical study," in *Proceedings of the 1988 connectionist models summer school*, 1988.

[17] P. Haffner, A. Waibel, H. Sawai, and K. Shikano, "Fast back-propagation learning methods for large phonemic neural networks," in *Proc. of the Eurospeech-89*, 1989.

[18] F. K. Soong and E. Huang, "A tree-trellis based fast search for finding the n-best sentence hypotheses in continuous speech recognition," in *Proceedings of the third DARPA workshop on speech and natural language*, Los Altos, CA, 1990, Morgan Kaufmann Publishers, Inc.

[19] K. Kita, T. Kawabata, and H. Saito, "HMM continuous speech recognition using predictive LR parsing," in *Proceedings of the ICASSP-89*, 1989.

[20] T. Hanazawa, K. Kita, S. Nakamura, T. Kawabata, and K. Shikano, "ATR HMM-LR continuous speech recognition system," in *Proceedings of the ICASSP-90*, 1990.

[21] M. Tomita, *Efficient parsing for natural language - A fast algorithm for practical systems*, Kluwer Academic Publishers, 1986.

[22] C. Myers and L. Rabiner, "A level building dynamic time warping algorithm for connected word recognition," *IEEE Trans. on ASSP*, vol. 29, no. 2, pp. 284–279, 1981.